\documentclass[letterpaper, 10 pt, conference]{ieeeconf}  

\IEEEoverridecommandlockouts                              

\usepackage{graphics} 
\usepackage{epsfig} 
\usepackage{mathptmx} 
\usepackage{times} 
\usepackage{amsmath} 
\usepackage{amssymb}  
\usepackage{outlines}
\usepackage{xcolor}
\usepackage[ruled,vlined,linesnumbered]{algorithm2e}
\usepackage{gensymb}
\usepackage{caption}
\usepackage{subcaption}
\usepackage[colorlinks=true,linkcolor=black,anchorcolor=black,citecolor=black,filecolor=black,menucolor=black,runcolor=black,urlcolor=black]{hyperref}
\usepackage{siunitx}
\usepackage{newtxmath}  
\usepackage{mathtools}
\usepackage{cite}

\title{\LARGE \bf
PlaneSLAM: Plane-based LiDAR SLAM for Motion Planning in Structured 3D Environments
}

\author{Adam Dai$^{1}$, Greg Lund$^{2}$ and Grace Gao$^{2}$
\thanks{$^{1}$Department of Electrical Engineering,
        Stanford, CA 94305, USA,
        addai@stanford.edu}%
\thanks{$^{2}$Department of Aeronautics and Astronautics,
        Stanford, CA 94305, USA
        \{greglund, gracegao\}@stanford.edu}%
}

\newcommand{\R}{\ensuremath{\mathbb{R}}}

\DeclareMathOperator*{\minimize}{min.}

\DeclareMathOperator{\arctantwo}{arctan2}

\newcommand{\regtext}[1]{\mathrm{\textnormal{#1}}}

\newcommand{\SO}{\mathbf{SO}}

\newcommand{\normal}{\mathbf{n}}
\newcommand{\basis}{\mathbf{b}}
\newcommand{\ctr}{\mathbf{c}}

\newcommand{\residual}{\mathbf{r}}
\newcommand{\jacobian}{\mathbf{J}}

\newcommand{\trans}{\mathbf{t}}
\newcommand{\plane}{\mathcal{P}}

\newcommand{\mbf}[1]{{\mathbf{#1}}}
\newcommand{\transpose}{^\mathsf{T}}

\begin{document}

\maketitle
\thispagestyle{empty}
\pagestyle{empty}

\begin{abstract}

LiDAR sensors are a powerful tool for robot simultaneous localization and mapping (SLAM) in unknown environments, but the raw point clouds they produce are dense, computationally expensive to store, and unsuited for direct use by downstream autonomy tasks, such as motion planning.
For integration with motion planning, it is desirable for SLAM pipelines to generate lightweight geometric map representations.
Such representations are also particularly well-suited for man-made environments, which can often be viewed as a so-called ``Manhattan world" built on a Cartesian grid.
In this work we present a 3D LiDAR SLAM algorithm for Manhattan world environments which extracts planar features from point clouds to achieve lightweight, real-time localization and mapping.
Our approach generates plane-based maps which occupy significantly less memory than their point cloud equivalents, and are suited towards fast collision checking for motion planning.
By leveraging the Manhattan world assumption, we target extraction of orthogonal planes to generate maps which are more structured and organized than those of existing plane-based LiDAR SLAM approaches. 
We demonstrate our approach in the high-fidelity AirSim simulator and in real-world experiments with a ground rover equipped with a Velodyne LiDAR.
For both cases, we are able to generate high quality maps and trajectory estimates at a rate matching the sensor rate of 10 Hz.
\end{abstract}

\section{Introduction}

\begin{figure}
    \centering
    \begin{subfigure}[b]{0.45\textwidth}
        \centering
        \includegraphics[width=\textwidth]{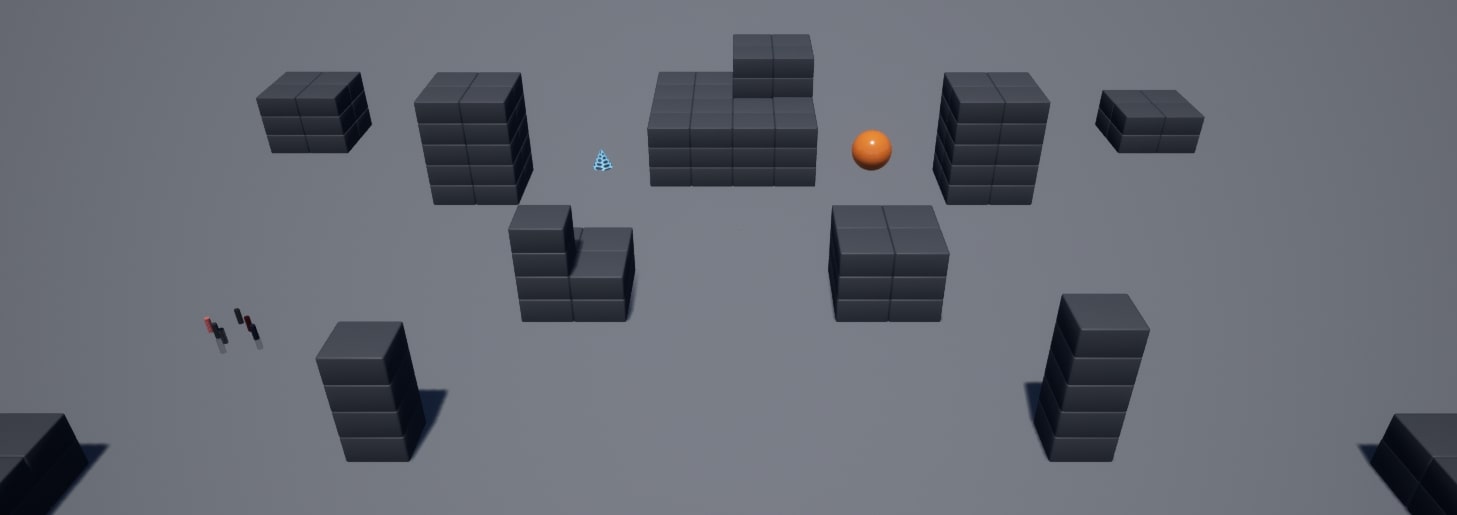}
        \caption{AirSim Blocks environment.}
        \label{fig:blocks}
    \end{subfigure}
    \hfill
    \begin{subfigure}[b]{0.47\textwidth}
        \centering
        \includegraphics[width=\textwidth]{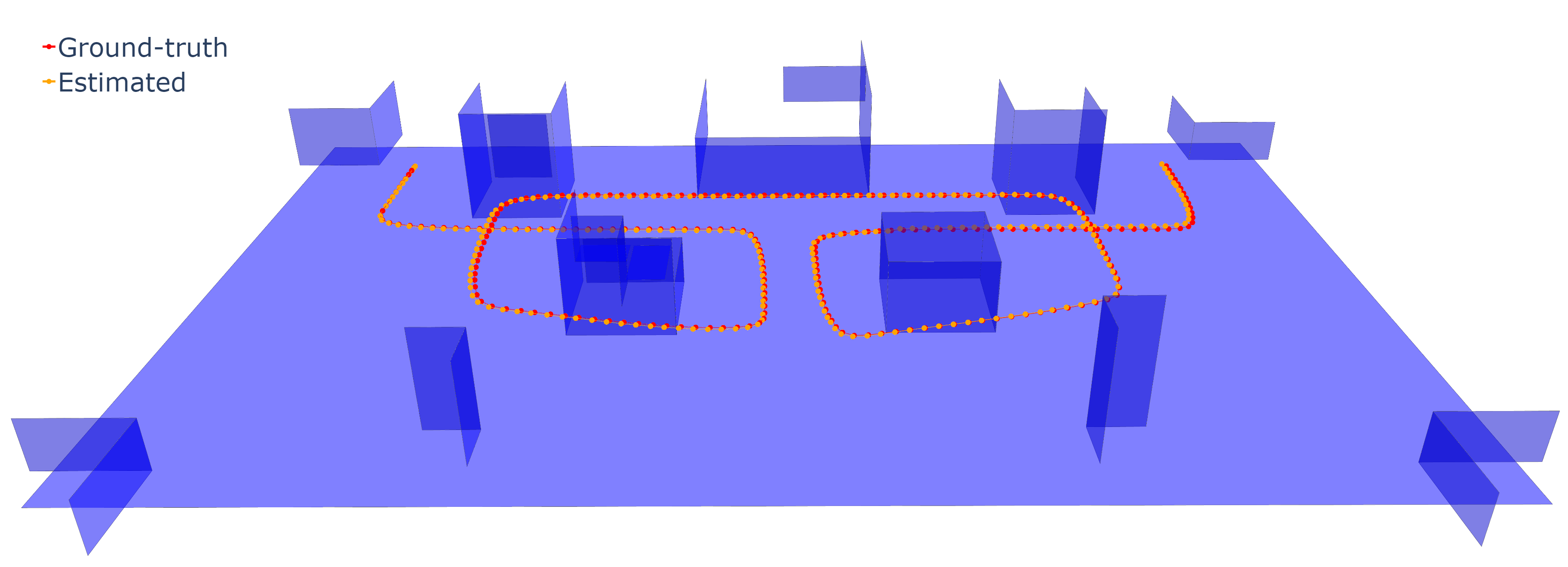}
        \caption{Map generated from our approach by drone flying through Blocks environment along the ground-truth trajectory shown in red. The trajectory estimated by our algorithm is shown in orange.}
        \label{fig:blocks_map}
    \end{subfigure}
    \caption{AirSim Blocks environment shown with corresponding plane map generated from our algorithm.}
    \label{fig:mapping}
\end{figure}

Simultaneous localization and mapping (SLAM) is a core functionality necessary for autonomous robots operating in unknown environments.
For robot navigation and obstacle avoidance, maps generated by SLAM may be used by downstream motion planning algorithms to plan collision-free paths through the environment, bridging the perception and planning layers of the autonomy stack.
Successful integration requires maps to be generated in real-time from sensor data, occupy minimal memory, and support fast and precise collision-checking.

LiDAR (Light Detection and Ranging) has emerged as a powerful sensor for spatial mapping and SLAM due to its ability to provide highly accurate depth information under a wide range of lighting and weather conditions.
However, operating directly on LiDAR point clouds can quickly become computationally expensive as the number of points increases, and point cloud maps created by LiDAR SLAM rapidly grow in memory as the robot explores more area.
One way to address this challenge is to leverage structure, and represent points which lie on the same plane with a single object, which saves memory and compute for later operations. 
This strategy is particularly effective for man-made urban and indoor environments, which consist primarily of mutually orthogonal planes.
These environments can be modeled as Manhattan worlds \cite{coughlan1999manhattan} and are conducive to lightweight geometric map representations.
Maintaining such minimal memory maps is especially important for embedded or resource-constrained robotic applications and for multi-robot settings, in which maps must be shared between robots over limited communication bandwidth.

Furthermore, converting point clouds to planar representations is beneficial for operations such as collision checking, which are essential to motion planning. 
Planes allow for continuous representation of boundaries in space, which can be described with concise mathematical expressions, whereas point clouds may visually form a surface, but mathematically are still only a discrete collection of points which, by themselves, do not define a boundary.
Traditionally, mapping and planning algorithms are designed separately, where maps produced by SLAM may not be compatible with motion planners, and planners may assume a map representation which is difficult to construct from real sensor data.
SLAM map representations which enable rapid collision checking allow for direct integration with downstream motion planning for autonomous mobile robots, and design of a fluid, real-time autonomy stack.

In this work, we present an approach for LiDAR-based SLAM in Manhattan worlds that extracts planes from point clouds for registration and map creation.
Contrary to most prior works in SLAM, our approach is designed with the task of motion planning in mind, as our geometric plane-based map representation enables fast and precise collision checking.
Furthermore, by registering sets of planes instead of dense point clouds, we speed up registration while maintaining accuracy, enabling faster odometry and loop closure. 
We leverage the Manhattan world assumption to extract orthogonal planes, and by merging these planes sequentially, we are able to create lightweight maps based on geometric primitives which capture the overarching 3D structure of the environment while occupying significantly less storage than their point cloud counterparts (an example such map is shown in Fig. \ref{fig:mapping}). 
Finally, we integrate our plane-based frontend with a pose graph backend for loop closure and global consistency, and demonstrate real-time localization and mapping, as well as rapid motion planning within our map with an RRT example.

\section{Related Work}

LiDAR SLAM has been well studied, and established methods such as LOAM~\cite{zhang2014loam} and its variations (LeGO-LOAM~\cite{shan2018lego}, F-LOAM~\cite{wang2021f}) rely on point clouds as a natural choice for registration and map representation.
These methods achieve accurate real-time localization and mapping, but the maps they produce are still point clouds, which grow rapidly in memory and are unsuited for direct use by motion planners.
Although point cloud maps are able to capture high levels of detail, for predominantly planar environments such as Manhattan worlds, they encode highly redundant information when compared to a direct plane representation.

Extracting planes from point clouds is not a new idea~\cite{yang2010plane, borrmann20113d, castagno2020polylidar3d}. 
One general strategy is to search for best-fit planes to the point cloud, which can be accomplished via RANSAC~\cite{yang2010plane, taguchi2013point} or the Hough transform~\cite{borrmann20113d, hulik2014continuous}.
Another strategy involves segmenting the point cloud into clusters of points which belong to the same plane, then extracting planes for each of these clusters.
Point cloud segmentation is commonly done using region growing~\cite{vo2015octree}, in which clusters of points are grown based on local surface normal similarity. 
These region growing methods rely on computing surface normals of local neighborhoods via principal component analysis (PCA). 
An alternative method for normal estimation is meshing the point cloud and computing normals for each triangle or polygon in the mesh.
Polylidar3D~\cite{castagno2020polylidar3d} introduced an algorithm for fast non-convex polygon extraction from depth sensor data which relies on triangulation and region growing.
One can also use meshes to construct lightweight scene representations, as in~\cite{rosinol2019incremental}.


Prior works have also explored plane-based registration and SLAM.
Pathak et al. \cite{pathak2010online} demonstrate plane-based registration and SLAM in real-world scenarios, generating maps that are orders of magnitude smaller than their point cloud equivalents.
However, their approach lacks solutions for merging planes together and is not suitable for real-time operation, as their core registration takes around 10 seconds. 
More recently, Grant et al. \cite{grant2019efficient} introduced Velodyne SLAM using point and plane features, extracting planes when possible to lower computation, while leveraging raw points when necessary to fully constrain the registration problem. 
Their approach generates maps consisting of polygonal planar patches and is shown to work well on a variety of indoor and outdoor environments, outperforming both Generalized ICP and LOAM, while running at rates between 2 and 10 Hz (depending on the number of points vs. planes used).
Favre et al. \cite{favre2021plane} also present an approach for plane-based point cloud registration which is shown to outperform traditional point-based methods while running at 1-3 Hz on various datasets, but does not consider the problem of SLAM and map generation.
Additionally, a variety of approaches for generating plane-based maps from dense depth information exist~\cite{taguchi2013point, salas2014dense, biswas2012planar}.

However, prior works in plane-based SLAM only present their maps for visualization, and do not consider the problem of motion planning within their plane maps. 
In contrast, our work highlights the advantage of plane maps for fast and efficient motion planning.
Additionally, through our plane extraction and merging approach, we are able to produce more structured and organized maps for Manhattan world environments.


\section{Problem Statement and Preliminaries}

\begin{figure}
    \centering
    \includegraphics[width=0.48\textwidth]{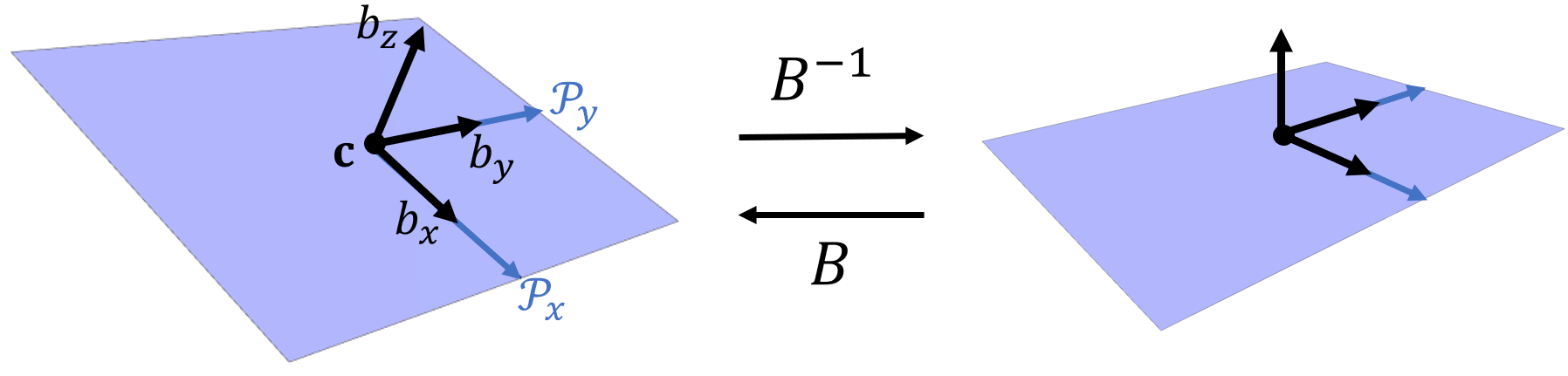}
    \caption{Example plane shown on the left parameterized by $\ctr$, $\plane_x$ and $\plane_y$ with basis $B = [\basis_x, \basis_y, \basis_z]$. By multiplying the plane parameters by $B^{-1}$, we map it to 2D such that it lies solely within the $xy$ plane.}
    \label{fig:basis}
\end{figure}

Given a robot equipped with a LiDAR sensor traveling through an unknown environment, our task is to accurately estimate the robot's trajectory while simultaneously constructing a sparse geometric map of the environment.
At frame $k$ our method takes as input an unorganized 3D point cloud $P_k = \{(x_1,y_1,z_1),\dots,(x_N,y_N,z_N)\}$.
We represent the trajectory as a sequence of poses $(R_k,\trans_k)$, where $R_k\in\R^{3\times3}$ is a rotation matrix describing orientation and $\trans_k\in \R^3$ is a translation vector describing position.
For our approach we represent the map as a collection of rectangular bounded planes (which we henceforth refer to simply as planes). 

We parameterize a plane $\plane$ with center point $\ctr$ and a pair of orthogonal vectors $\plane_x, \plane_y \in \R^3$ which span its area (see Fig. \ref{fig:basis}). 
We define a plane's basis as $B = [\basis_x, \basis_y, \basis_z]$, where $\basis_x = \plane_x/||\plane_x||$, $\basis_y = \plane_y/||\plane_y||$, and $\basis_z = \basis_x \times \basis_y$.
Note that $\basis_z$ is also the plane's normal vector $\normal$.
This basis can be used to map the plane to 2D, as shown in Fig. \ref{fig:basis}.
The 2D plane representation is essential to our plane extraction process, and for operations used later in plane merging and collision checking.

\section{Approach}

%

\begin{figure}
    \centering
    \includegraphics[width=0.47\textwidth]{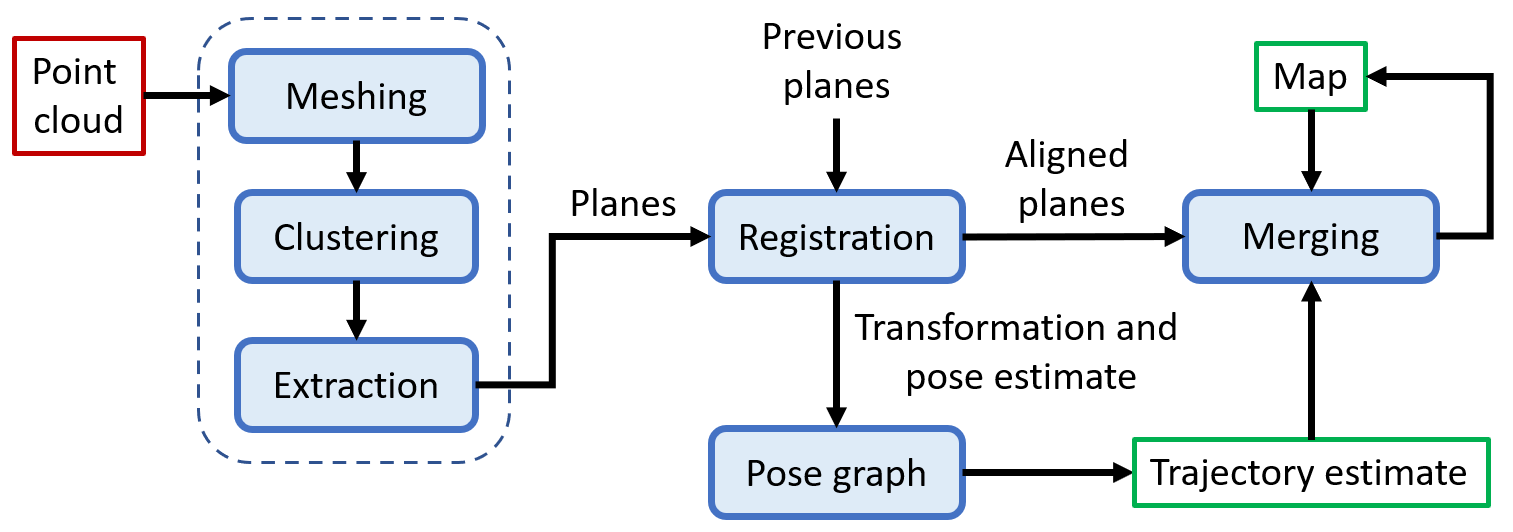}
    \caption{Approach overview (inputs are boxed in red and outputs are boxed in green).
    First planes are extracted from the input point cloud, and these planes are then registered with previous planes for odometry. A pose graph is maintained using the odometry estimates, and map maintained and updated by merging newly detected planes in.}
    \label{fig:overview}
\end{figure}

As with most SLAM pipelines, our approach consists of a frontend which performs feature extraction and registration, and a backend which maintains and optimizes a pose graph and handles map generation. 
Our approach overview is shown in Fig. \ref{fig:overview}. 
We now describe the components of our approach in detail.

\subsection{Plane Extraction}

When we receive a new point cloud from our LiDAR sensor (Fig. \ref{fig:frontend_points}), the first step is to extract planar features from it. 
Our extraction approach uses region growing for point cloud clustering, with meshing for surface normal estimation and graph structure.


\subsubsection{Meshing}


In order to mesh the point cloud, we first reparameterize each point in spherical coordinates relative to the scan-centered coordinate frame:
\begin{align}
\theta_i = \arctantwo(y_i,x_i), \ \phi_i = \arctantwo(z_i, \sqrt{x_i^2 + y_i^2}). 
\end{align}
We then compute the Delaunay triangulation \cite{berg1997computational} of the $(\theta_i, \phi_i)$ points. 
The result of the triangulation is a set of simplices $(i,j,k)$, where $i,j,k \in \{1,\dots,N\}$ are indices of points which form a triangle in the triangulation.
Lastly, we prune the mesh to remove triangles with side lengths which exceed a certain threshold and smooth the mesh with a Laplacian filter \cite{field1988laplacian} to reduce noise.
Fig. \ref{fig:frontend_mesh} shows an example of point cloud meshing.

\subsubsection{Clustering}

Next, we cluster the triangles in the mesh according to their surface normal and locality.
We do this through breadth-first graph search over the mesh triangles.
First, we compute a surface normal vector for each triangle in the mesh by taking the cross product of two of its edge vectors $\normal = \mathbf{e}_1 \times \mathbf{e}_2$. 
We then randomly select a root triangle to begin the cluster and initialize a search queue $S$ with its neighbors.
At each iteration, a new triangle is removed from $S$, and its normal vector is compared to the root triangle's normal by taking their dot product.
If the normals are close enough (i.e., their dot product exceeds a threshold), the triangle is added to the current cluster, and its neighbors are added to $S$.
The cluster is grown until $S$ becomes empty, indicating there are no more neighboring triangles with normals similar to the root's normal.
This process is repeated with a new unclustered root triangle until all triangles in the mesh have been clustered.
The mesh clustering process is summarized in Algorithm \ref{alg:clustering}.

\begin{algorithm}
\SetAlgoLined
\KwIn{Mesh $M$, normals $\normal$}
\KwOut{Clusters $C$}
\SetKwFunction{pop}{pop}
\SetKwFunction{append}{append}
\SetKwFunction{remove}{remove}
\SetKwFunction{update}{update}
\SetKwFunction{len}{len}
 
 $Q$ = \{set of all Mesh indices\}\;
 $C$ = []\;
 
 \While{$Q$ not empty}{
    root = $Q$.\pop{}\;
    cluster = [root]\;
    $S$ = \{neighbors of root also in $Q$\}\;
    \While {$S$ not empty}{
        $i$ = $S$.\pop{}\;
        \If{$\normal[i]\transpose\normal[\regtext{root}] > thresh$}{
            cluster.\append{i}\;
            $Q$.\remove{i}\;
            $S$.\update{\textnormal{neighbors of $i$ also in $Q$}}\;
        }
    }
    $C$.\append{\textnormal{cluster}}\;
 }
 \caption{Mesh clustering}
\label{alg:clustering}
\end{algorithm}

\begin{figure}
\centering
    \begin{subfigure}[b]{0.23\textwidth}
        \centering
        \includegraphics[width=\textwidth]{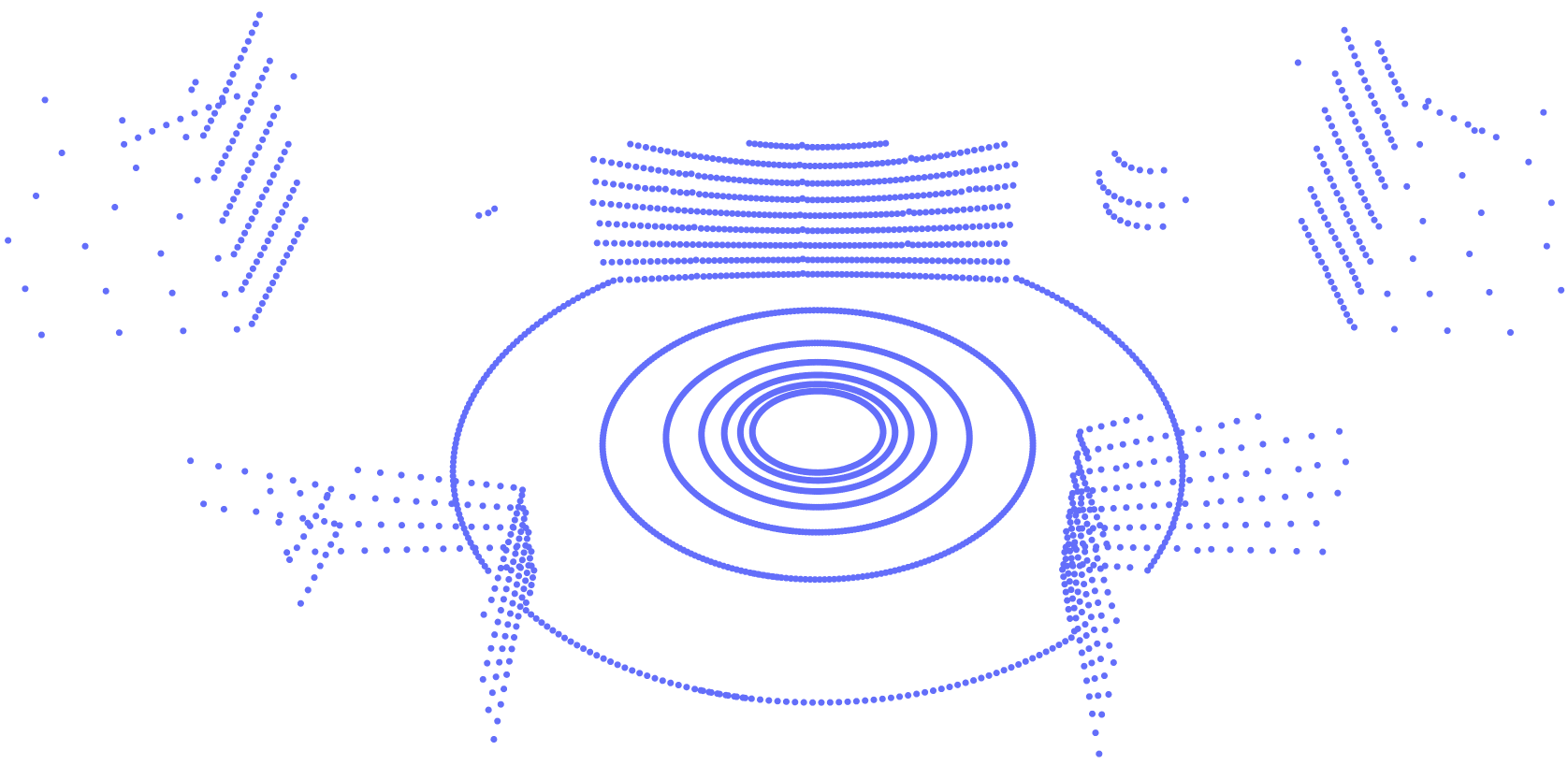}
        \caption{Points}
        \label{fig:frontend_points}
    \end{subfigure}
    \hfill
    \begin{subfigure}[b]{0.23\textwidth}
        \centering
        \includegraphics[width=\textwidth]{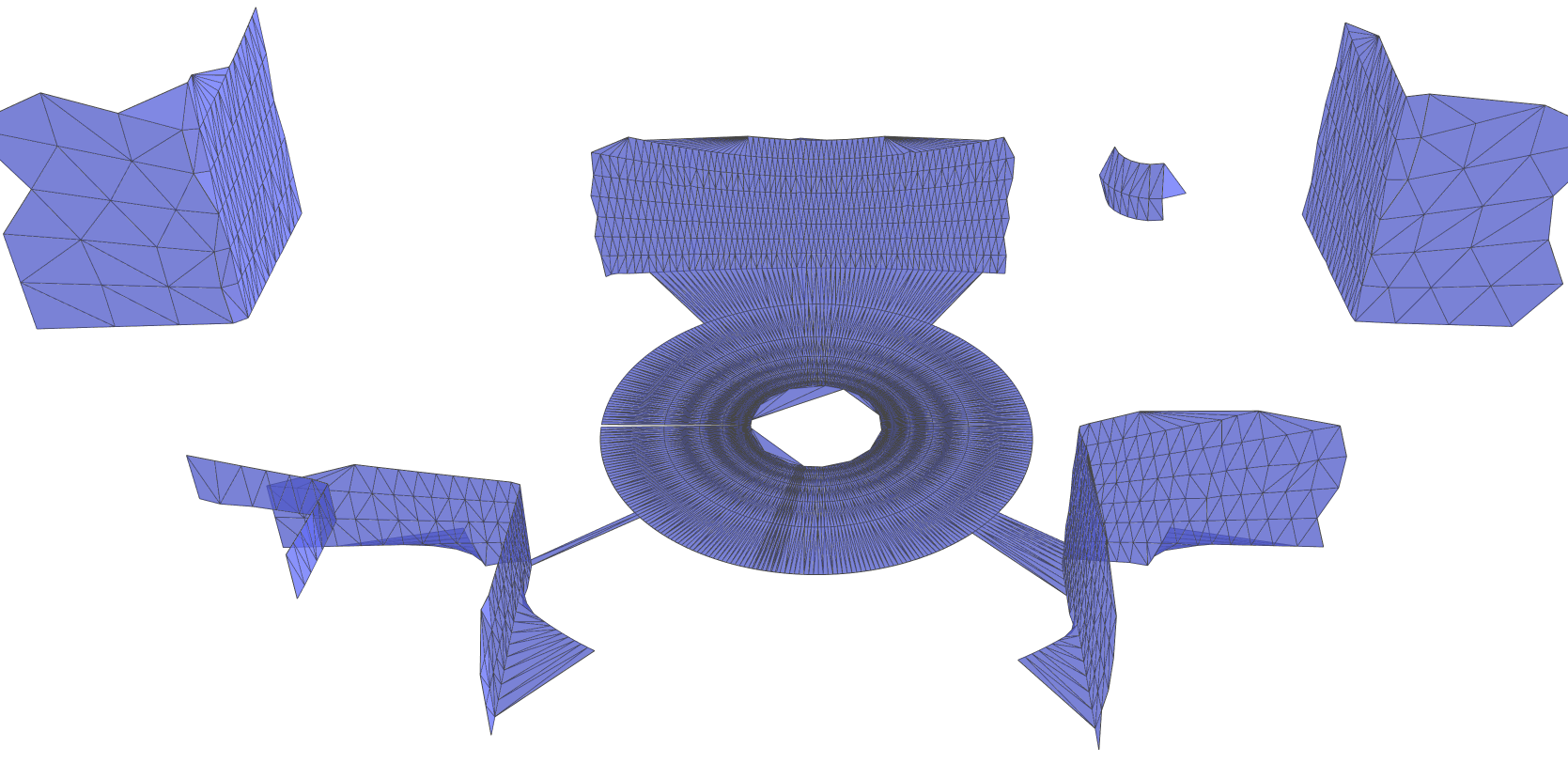}
        \caption{Mesh}
        \label{fig:frontend_mesh}
    \end{subfigure}
    \hfill
    \begin{subfigure}[b]{0.23\textwidth}
        \centering
        \includegraphics[width=\textwidth]{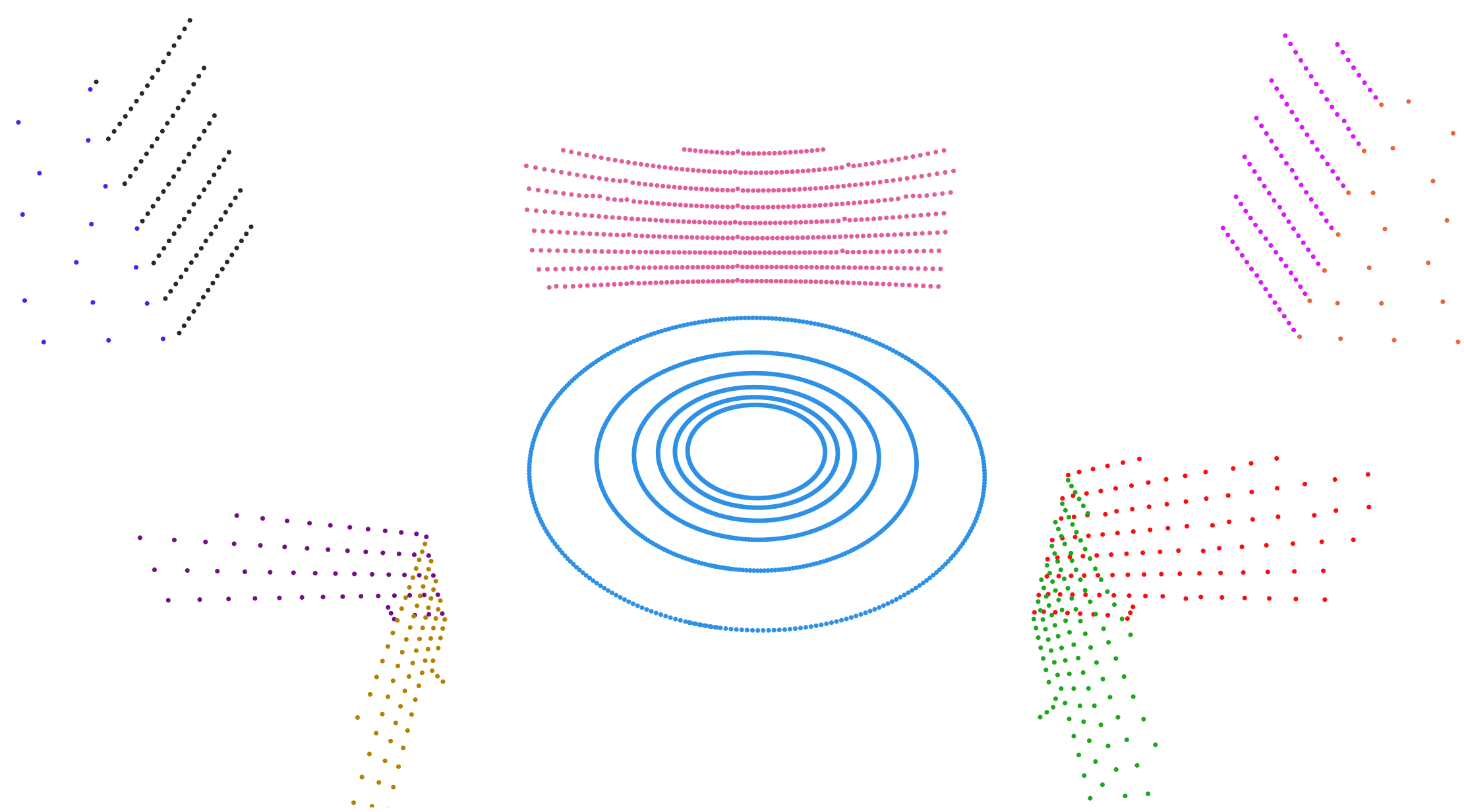}
        \caption{Clusters}
        \label{fig:frontend_clusters}
    \end{subfigure}
    \hfill
    \begin{subfigure}[b]{0.23\textwidth}
        \centering
        \includegraphics[width=\textwidth]{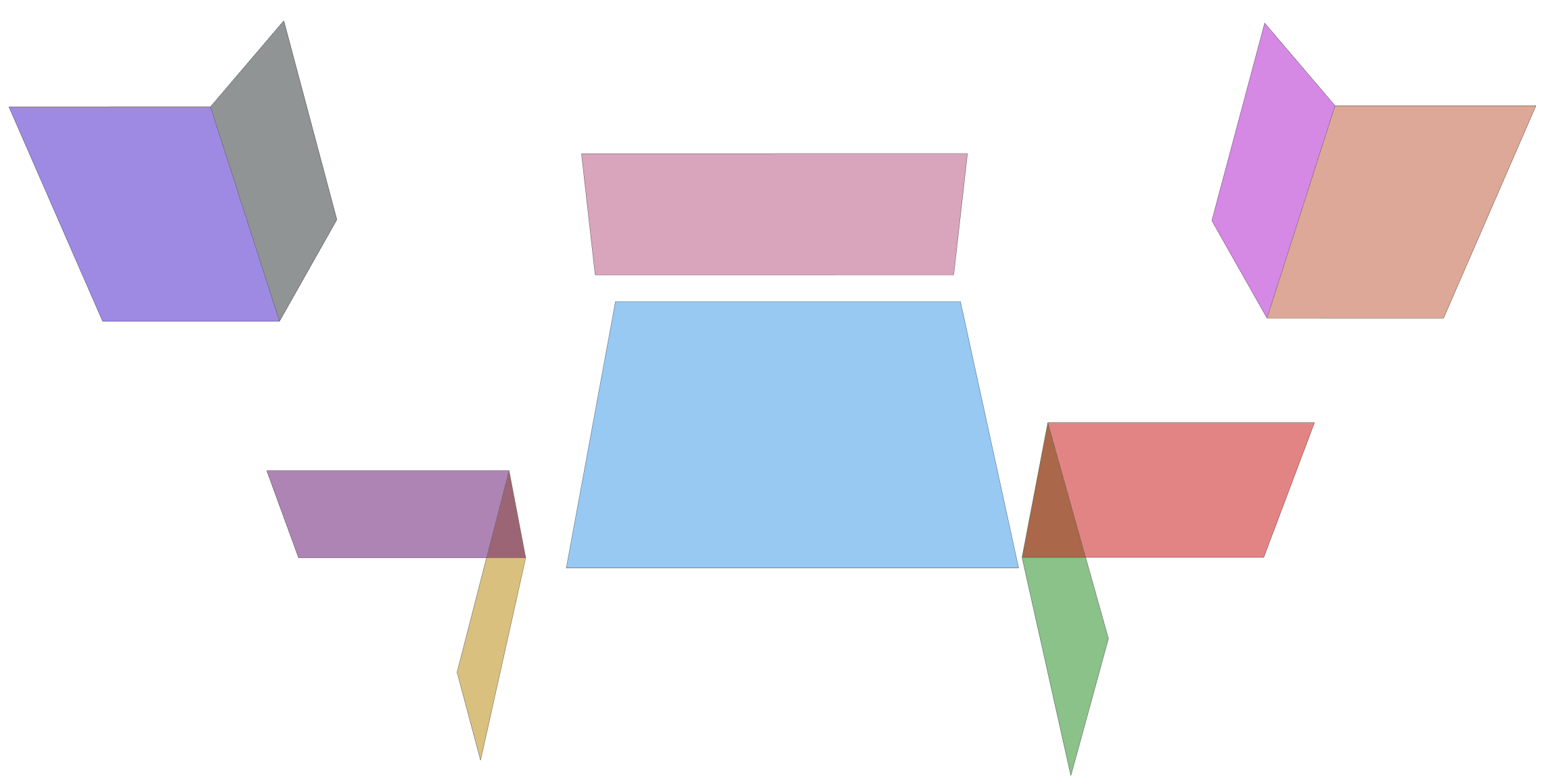}
        \caption{Planes}
        \label{fig:frontend_planes}
    \end{subfigure}
\caption{Plane extraction process. First the point cloud (\ref{fig:frontend_points}) is triangulated to form a mesh (\ref{fig:frontend_mesh}). Then graph search is used to cluster the mesh triangles and corresponding points (\ref{fig:frontend_clusters}). Finally, planes (\ref{fig:frontend_planes}) are extracted for each cluster of points.}
\label{fig:frontend}
\end{figure}

Once we have a clustering of the mesh triangles, we can use it to extract a clustering over the points themselves. 
Fig. \ref{fig:frontend_clusters} shows an example of clustered points.

\subsubsection{Extraction}


Given clusters of points $\{P^{(1)},\dots,P^{(k)}\}$ with associated average normals $\{\normal^{(1)},\dots,\normal^{(k)}\}$ (computed as the normalized average of the normal vectors of all the mesh triangles in the cluster), we now aim to extract a plane for each cluster.
First we determine a basis $B = [\basis_x, \basis_y, \basis_z]$ for the ground plane.
Then, leveraging the Manhattan world assumption, all planes are extracted such that their normals aligns with either $\basis_x, \basis_y$ or $\basis_z$, and the resulting set of extracted planes are mutually orthogonal.


To find $B$, we first identify the ground plane as the largest cluster with normal closest to the $z$ direction (i.e., $(0,0,1)$), and set its normal to $\basis_z$.
Next, the largest cluster with normal approximately orthogonal to the ground plane is found, and $\basis_x$ is chosen as the orthogonal projection of its normal onto the ground plane. 
Finally $\basis_y = \basis_z \times \basis_x$.

Once we have the extraction basis, the plane extraction procedure for each cluster $P^{(i)}$ (with average normal $\normal^{(i)}$) is as follows.
First we project $P^{(i)}$ onto $B$ to obtain $P^{(i)}_B =~B^{-1}P^{(i)}$.
Next we find the basis vector closest to $\normal^{(i)}$ and extract a 2D bounding box over the points in the remaining dimensions (e.g. if $\normal^{(i)}$ is closest to $\basis_x$, then the bounding box is extracted over $y$ and $z$).
Once the 2D bounding box over $P^{(i)}_B$ is found, the center point and the vectors which span the bounding box are projected back to the standard basis to obtain the parameters $\ctr, \plane_x$ and $\plane_y$ for the final extracted plane.

The output of the extraction is a plane set $\plane^k =~\{\plane^k_1,\dots,\plane^k_n\}$ (example shown in Fig. \ref{fig:frontend_planes}). For plane $\plane^k_i$, $\normal^k_i$ is its normal vector, $\ctr^k_i$ is its center point, and $d^k_i$ is its distance to origin (computed as $d^k_i = \normal^k_i{\transpose} \ctr^k_i$).

\subsection{Plane Registration}

Next, in order to perform odometry and loop closure, we must be able to perform registration between two plane sets.
Given a source plane set $\plane^s$ and target plane set $\plane^t$ with known correspondences denoted by the common index $i$, we formulate the registration problem as
\begin{align} \begin{split} \label{prob:registration}
    \minimize_{R, \trans} \quad  & \sum_{i=1}^n ||R\normal^s_i - \normal^t_i||_2^2 + ||(R\normal^s_i)\transpose \trans + d^s_i - d^t_i||_2^2 \\
    \regtext{s.t.} \quad  & R\in \SO(3), \ \trans\in \R^3
\end{split} \end{align}


\subsubsection{Correspondences} \label{subsec:correspondences}
In order to determine correspondences, we compute a similarity cost metric for each pair of source and target planes based on their normal vectors and center-to-center distance
\begin{align}
    c(\plane^s_i,\plane^t_j) = \alpha ||\normal^s_i - \normal^t_j|| + \beta ||\ctr^s_i - \ctr^t_j||
\end{align}
where $\alpha$ and $\beta$ are user-specified scaling parameters. 
We then form a cost matrix from the $c(\plane^s_i,\plane^t_j)$, and select the minimum value from each row (i.e., lowest cost assignment for each source plane to a target plane). 
Additionally, if different source planes are assigned to the same target plane, we choose only the lowest cost assignment.
In this way, each plane in both the source and target is uniquely corresponded, which we observed to reduce false correspondences.

\subsubsection{Registration}

As shown in \cite{pathak2010online, taguchi2013point}, the rotation and translation components of \eqref{prob:registration} can be decoupled.
For rotation estimation, we define the rotation residual $r_i(R) = R\normal^s_i - \normal^t_i$ and use the Gauss-Newton method to solve
\begin{align} \begin{split}
    \minimize_{R} \quad  & \sum_i^n ||r_i(R)||_2^2 \\
    \regtext{s.t.} \quad  & R\in \SO(3).
\end{split} \end{align}
Since $R \in \SO(3)$, we use the Lie algebra parameterization~\cite{blanco2010tutorial} to map the problem to an unconstrained optimization over $\R^3$. 
The Gauss-Newton update proceeds as 
\begin{subequations}
\begin{align}
    r &= \residual(R^{(i)}) \\
    J &= \jacobian(R^{(i)}) \\
    d\omega &= -\mu (J^T J + \lambda I)^{-1} J^T r \\
    R^{(i+1)} &= \exp(d\omega) R^{(i)}
\end{align}
\end{subequations}
where $\mu$ and $\lambda$ are the update step and regularization parameters respectively, $\exp$ is the $\SO(3)$ exponential map, $\residual$ is the stacked rotation residual, and $\jacobian$ is the Jacobian of $\residual$ 
\begin{align}
    \residual(R) = \begin{bmatrix} r_1(R) \\ \vdots \\ r_n(R) \end{bmatrix}, \
    \jacobian(R) = \begin{bmatrix}  -(R \normal^s_1)^\wedge \\ \vdots \\ -(R \normal^s_n)^\wedge \end{bmatrix}
\end{align}
where $^\wedge$ denotes the skew symmetric operator.  
The derivation of $\jacobian$ can be achieved using chain rule with Equation 10.8 from \cite{blanco2010tutorial}.

After we obtain a rotation estimate $\hat{R}$ from Gauss-Newton, we set $R = \hat{R}$ and solve for $\trans$
\begin{align} \begin{split}
    \minimize_{\trans} \quad  & \sum_i^n ||(\hat{R}\normal^s_i)\transpose \trans + d^s_i - d^t_i||_2^2 \\
    \regtext{s.t.} \quad  & \trans \in \R^3
\end{split} \end{align}
We can formulate this as a least squares problem with
\begin{align}
    A = \begin{bmatrix} \hat{R}\normal^s_1 \\ \vdots \\ \hat{R}\normal^s_n \end{bmatrix}, \
    b = \begin{bmatrix} d^t_1 - d^s_1 \\ \vdots \\ d^t_n - d^s_n \end{bmatrix}.
\end{align}
The least squares solution is thus given by $\hat{\trans} = (A\transpose A)^{-1} A\transpose b$.
Note that the least squares problem is rank deficient unless there are at least three correspondences whose normals span $\R^3$. 
Our current implementation assumes this condition is always satisfied, although this issue has been addressed in priors works \cite{pathak2010online}.

\subsubsection{Outlier Correspondence Removal}

As the computed correspondences are not always perfect, there is a need to make our registration robust to faulty correspondences. We employ an outlier removal method inspired by residual-based fault detection and exclusion \cite{parkinson1988autonomous}, in which first the magnitude of the translation residual is computed to determine if a fault is present. If so, then the correspondence associated with the largest component of the residual is iteratively removed until the optimization yields a solution under the desired threshold. 

\subsection{Pose Graph}



We employ the standard pose-graph formulation presented in \cite{grisetti2010tutorial}, in which nodes of the graph represent the pose of the robot at different time frames, and edges between nodes represent measurements.
The pose graph is initialized with some initial pose $(R_0, \trans_0)$.
At frame $k$, a new LiDAR scan is taken and plane set $\plane^k$ is extracted, which is registered with $\plane^{k-1}$ to compute relative transformation $(R_{k-1}^k, \trans_{k-1}^k)$. 
This transformation is used to compute the new absolute pose estimate for current frame $k$:
\begin{align} \begin{split}
    R_k &= R_{k-1}^k R_{k-1} \\ 
    \trans_k &= \trans_{k-1} + R_{k-1} \trans_{k-1}^k
\end{split} \end{align}
which is then used to initialize the new node $k$, and $(R_{k-1}^k, \trans_{k-1}^k)$ is also used to create an odometry edge between node $k-1$ and $k$.

After node $k$ is added to the graph, we search nearby nodes for loop closure.
If the distance between nonconsecutive nodes $k$ and $j$ is below our search radius, their plane sets are registered to compute the relative transformation for a loop closure edge.
Since in loop closure situations, scans can have large relative transformations, the absolute poses of the nodes are used to first transform $\plane^k$ and $\plane^j$ to the world frame to obtain correspondences, before the registration is run on the original planes in the local frame.
After a loop closure edge is added, the pose graph is optimized as in \cite{grisetti2010tutorial}.

\subsection{Plane Set Merging}

Our map is formed through a recursive merging process, in which the first plane set is initialized as the map, and successive plane sets are merged in. 
Once a plane set has been registered and its absolute pose estimated, it is transformed and merged with the current version of the map.
The merging process takes as input two plane sets $\plane^s$ and $\plane^t$, and iterates through the planes in $\plane^s$.
For each plane $\plane^s_i \in \plane^s$, we then iterate through each plane $\plane^t_j \in \plane^t$ and check three conditions:
\begin{enumerate}
    \item The two planes are approximately coplanar. 
    \item The plane-to-plane distance is below a threshold.
    \item The 2D projections of $\plane^s_i$ and $\plane^t_j$ onto $\plane^s_i$'s basis overlap.
\end{enumerate}
If these conditions are satisfied, then $\plane^t_j$ is added to $\plane^s_i$'s correspondence set. 
After we have iterated through all planes in $\plane^t$, the corresponding planes are projected onto $\plane^s_i$'s basis, and the 2D bounding box over $\plane^s_i$ and all its corresponding planes is found. 
This box is then projected back to the standard basis to obtain the new merged plane.
In this way, all corresponding planes in $\plane^s$ and $\plane^t$ are merged to form a new overall merged plane set.

Following merging, we also clean up the map by removing planes with small area and fusing the edges of planes near each other.
In the event of loop closure and pose graph optimization, the map must be regenerated by transforming all plane sets $\plane^0,\dots,\plane^k$ based on the newly re-optimized trajectory and re-merging them all together.

\subsection{Collision Checking}\label{subsec:collision_checking}

We briefly describe a fast collision check procedure for a plane and line segment. 
Although we focus on this specific example, which is used to generate an RRT later in our results, the plane representation also supports efficient collision checking with other types of trajectory representations and parameterizations.

Given a line segment $\ell$ represented by two points $\mbf{a}, \mbf{b} \in~\R^3$ and plane $\plane$ with normal $\normal$ and center $\ctr$, we wish to determine if $\ell$ intersects with $\plane$.
First we project both $\ell$ and $\plane$ to $\plane$'s basis.
Since the remainder of the intersection check process deals only with projected versions of $\ell$ and $\plane$, we will hereby refer to the projected versions as $\ell$ and $\plane$.

Next we compute the intersection of the infinite line formed by extending $\ell$ with the infinite plane that $\plane$ lies in. 
The infinite plane can be represented as $\{\mbf{x}\ |\ \normal\transpose \mbf{x} = \ctr\}$, and the infinite line can be represented as $\{c \mbf{v} + \mbf{a}\ |\ c\in\R\}$ where $\mbf{v} = \mbf{b} - \mbf{a}$.
Thus we can find the intersection by solving for $c\in\R$ such that $\normal\transpose (c\mbf{v} + \mbf{a}) = \ctr$.
Furthermore, since we have projected to $\plane$'s basis, we know that the projected plane lies in the $xy$ plane, and $\normal = (0,0,1)$.
Thus we have $c v_3 + a_3 = {\ctr}_3$ so $c = ({\ctr}_3 - a_3) / v_3$.
If $v_3 = 0$ then $\ell$ is parallel to the $xy$ plane and there is no intersection.
Otherwise, we then proceed to check if the intersection point $c\mbf{v} + \mbf{a}$ lies within the line segment and plane boundaries.
To check the line segment constraint, we simply check if $0 \le c \le 1$.
To check the plane constraint, we check if the 2D box defined by $xy$ dimensions of $p$ contains $c\mbf{v} + \mbf{a}$.
If both of these conditions are satisfied, then $\ell$ and $\plane$ intersect, otherwise, they do not intersect.
\section{Results}

We evaluate our algorithm both in simulation and on hardware. 
For simulation, we use AirSim \cite{shah2018airsim} to simulate a drone flying through an environment and collecting LiDAR measurements. 
For our hardware tests, we use a ground rover equipped with a Velodyne Puck LITE LiDAR shown in Fig. \ref{fig:flightroom}. 
Our method is implemented in python with documented code available online\footnote{\url{https://github.com/Stanford-NavLab/planeslam}}.
The python-graphslam library \cite{irion2019pythongraphslam} is used for pose graph creation and optimization.

\subsection{AirSim Experiments}

We use the ``Blocks" environment, shown in Fig. \ref{fig:blocks}, which consists of several rectangular blocks placed in an open space. 
The AirSim LiDAR is configured with 16 channels, 60$\degree$ vertical field of view (FOV), 360$\degree$ horizontal FOV, 10 rotations per second, and 100,000 points per second, and the experiments are run on a 6-core, 3.6 GHz desktop with 32~GB RAM.


We show results for a trajectory in which the drone traverses the central area of the environment while LiDAR point clouds are collected at 10 Hz. 
Fig. \ref{fig:mapping} shows the final map generated by our approach along with the ground-truth and estimated trajectories.
For trajectory of total length \SI{603.6}{\m}, the RMSE translational error is \SI{1.3}{\m} with standard deviation (std.) \SI{0.6}{\m}, and the average rotational error (computed as the angle between the estimated and ground-truth rotational frames) is 0.30 degrees with std. 0.14 degrees. 


Table \ref{tab:timing} shows a breakdown of the average runtime of each module of our method in milliseconds.
Note that loop closure has a high standard deviation as it is not run every iteration.
Our overall algorithm has an average runtime of \SI{88.8}{\ms} with standard deviation \SI{24.6}{\ms}, which is comparable to LeGO-LOAM and F-LOAM, and is fast enough to be run at the LiDAR rate of 10 Hz.
\begin{table}
    \centering
    \begin{tabular}{l c c c}\hline \noalign{\vskip 1mm}   
        \textbf{Module} & \textbf{Average (ms)} & \textbf{Standard Deviation (ms)} \\ \noalign{\vskip 1mm} \hline    
        \noalign{\vskip 1mm}
        Extraction     & 43.5 & 6.8 \\
        Registration   & 7.2  & 3.2 \\
        Loop closure   & 15.6 & 46.9 \\
        Merging        & 22.3 & 12.6 \\\hline \noalign{\vskip 0.5mm}   
        \textbf{Total} & 88.8 & 24.6 \\\hline
    \end{tabular}
    \caption{Average runtimes per iteration for each module of our method. Our method can be run in real-time at 10 Hz and is comparable in speed to state-of-the-art approaches.}
    \label{tab:timing}
\end{table}

We also compare the size of our plane map in Fig. \ref{fig:blocks_map} with the size of an equivalent point cloud map.
The plane map occupies 41.9 kB of memory, whereas a point cloud map generated from superimposing every 10th point cloud together occupies 34.5 MB of memory.
Thus we see that by using a plane map representation, we are able to reduce memory usage by approximately three orders of magnitude.

\begin{figure}
    \centering
    \includegraphics[width=0.5\textwidth]{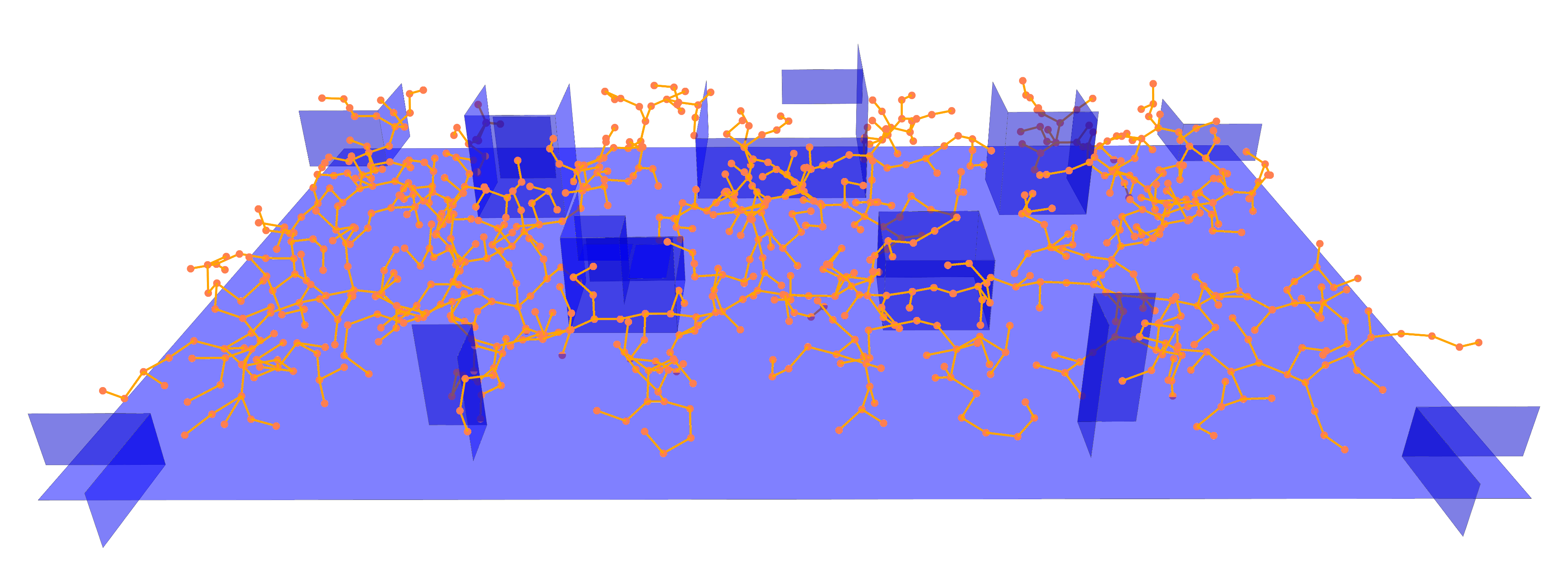}
    \caption{RRT generated in \SI{0.5}{\s}, demonstrating the speed of collision checking within our plane-based map.}
    \label{fig:rrt}
\end{figure}

In order to demonstrate the suitability of our plane map for motion planning, we implement a naive straight-line RRT~\cite{lavalle1998rapidly} using the collision check process described in Section~\ref{subsec:collision_checking}.
An RRT with 1000 nodes is shown in Fig. \ref{fig:rrt}, and takes only \SI{0.5}{\s} to generate.

\subsection{Rover Experiments}

\begin{figure}
    \centering
    \begin{subfigure}[b]{0.4\textwidth}
        \centering
        \includegraphics[width=\textwidth]{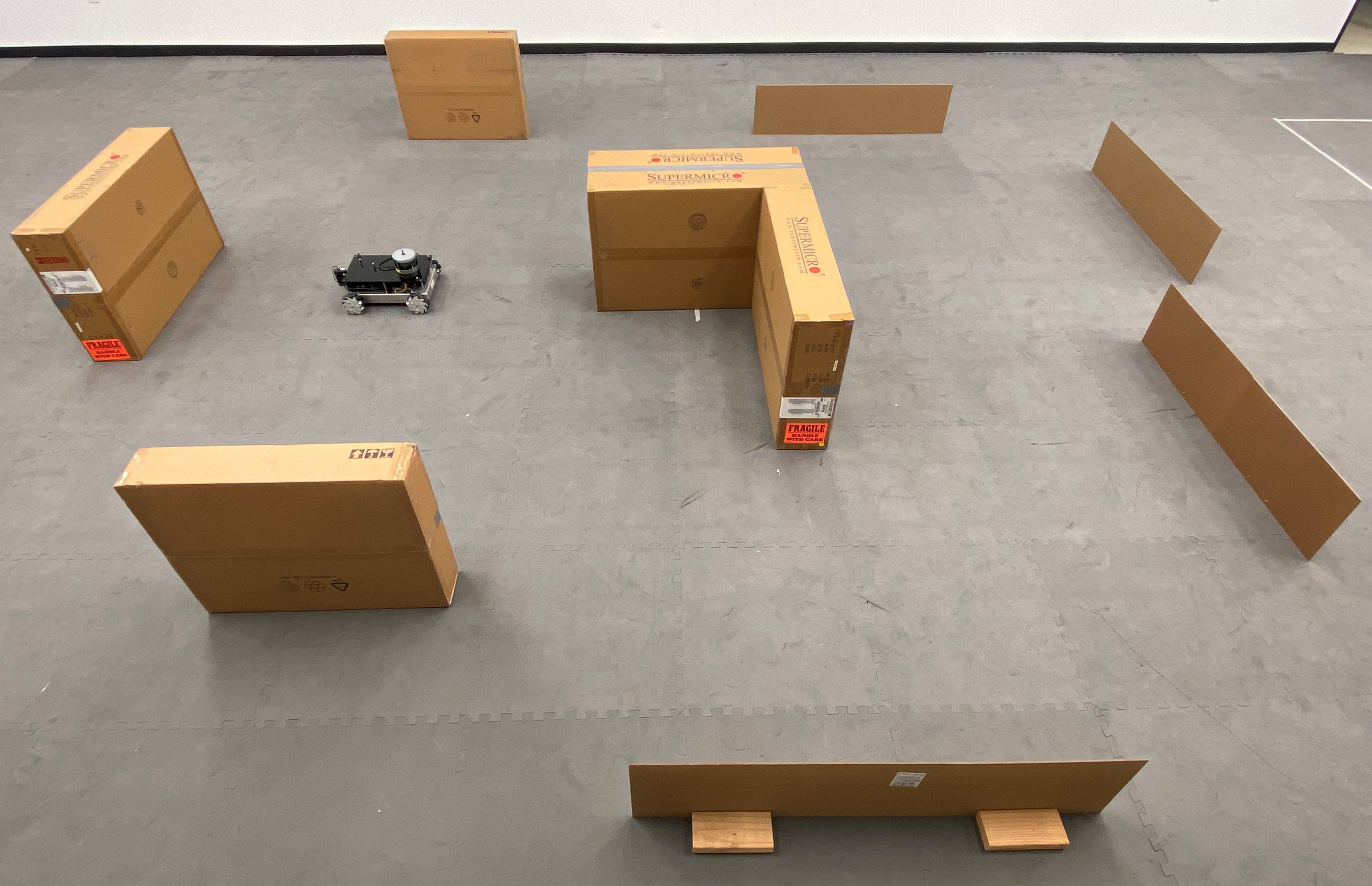}
        \caption{Ground rover in cardboard box environment.}
        \label{fig:flightroom}
    \end{subfigure}
    \hfill
    \begin{subfigure}[b]{0.48\textwidth}
        \centering
        \includegraphics[width=\textwidth]{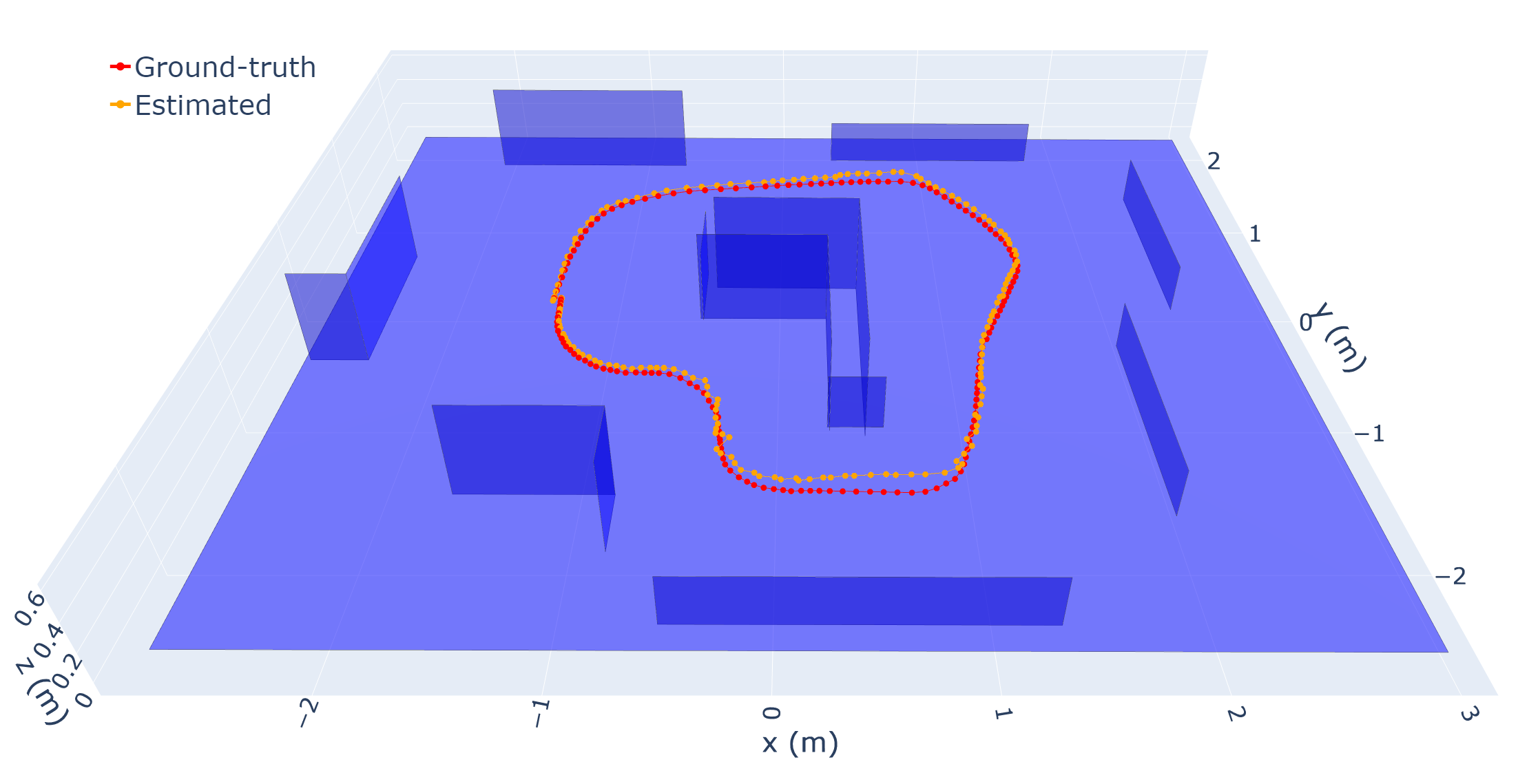}
        \caption{Generated plane map for cardboard box environment with ground-truth trajectory in red and estimated trajectory in orange.}
        \label{fig:rover_map}
    \end{subfigure}
    \caption{We validate our method with a ground rover equipped with Velodyne LiDAR in a Manhattan world environment. Our method is able to estimate the rover's trajectory and construct a plane-based map of the environment in real-time.}
    \label{fig:rover_mapping}
\end{figure}

For our hardware tests, we use a ground rover equipped with a Velodyne Puck LITE LiDAR, which has 16 channels, 30$\degree$ vertical FOV, 360$\degree$ horizontal FOV, and generates 300,000 points per second. 
We manually drive the rover through an environment made up of cardboard boxes, shown in Fig. \ref{fig:flightroom}, and ground-truth pose information is logged using motion capture.
The estimated map and trajectory from our algorithm are shown in Fig. \ref{fig:rover_map}.
The RMSE translational error for the trajectory is \SI{0.063}{\m} (std. \SI{0.039}{\m}), and the average rotational error is 0.71 degrees (std. 0.40 degrees).

All computation is done onboard the rover on an Intel NUC 7 mini PC (2-core, 3.5 GHz, 16 GB RAM), with total average runtime of \SI{124}{\ms} (std. \SI{37}{\ms}).
Although the Velodyne LiDAR produces scans at 10 Hz, we run our algorithm at 5 Hz (every other scan) to ensure sufficient compute time available.

We observe that for real-world noisy LiDAR data, the frontend is susceptible to error, with some planes being slightly over-approximated. 
Additionally, errors in the trajectory and map can be contributed to the rotation of the LiDAR during the rover's motion, which is not accounted for.
Nevertheless, our method is still able to reconstruct the overall structure of the cardboard box environment.


\section{Conclusion}

In this paper, we present plane-based LiDAR SLAM for real-time lightweight map generation and localization.
Traditionally, SLAM algorithms are not designed with the requirements of downstream motion planning algorithms in mind, and motion planning algorithms are often designed assuming a map of the environment exists in a convenient form.
Our work takes a step towards bridging this gap, proposing a SLAM pipeline that can be tightly integrated with motion planning through its plane-based map representation. 
Additionally, our map requires much less memory to store than a point cloud equivalent, enabling applications involving extensive mapping on resource-constrained systems, or multi-robot applications in which robots must communicate map information between each other over limited bandwidth.

\bibliographystyle{IEEEtran}
\bibliography{references}

\end{document}